# Learning Structured Twin-Incoherent Twin-Projective Latent Dictionary Pairs for Classification

Zhao Zhang[1,2,3], Yulin Sun[1], Zheng Zhang[4], Yang Wang[2,3], Guangcan Liu[5] and Meng Wang[2,3]

[1] School of Computer Science and Technology, Soochow University, Suzhou 215006, China
[2] Key Laboratory of Knowledge Engineering with Big Data (Ministry of Education), Hefei University of Technology
[3] School of Computer Science and Information Engineering, Hefei University of Technology, Hefei, China
[4] Bio-Computing Research Center, Harbin Institute of Technology (Shenzhen), Shenzhen, China
[5] School of Information and Control, Nanjing University of Information Science and Technology, Nanjing, China
e-mail: cszzhang@gmail.com, daitusun@gmail.com

*Abstract*— In this paper, we extend the popular dictionary pair learning (DPL) into the scenario of twin-projective latent flexible DPL under a structured twin-incoherence. Technically, a novel framework called Twin-Projective Latent Flexible DPL (TP-DPL) is proposed, which minimizes the twin-incoherence constrained flexibly-relaxed reconstruction error to avoid the possible over-fitting issue and produce accurate reconstruction. In this setting, our TP-DPL integrates the twin-incoherence based latent flexible DPL and the joint embedding of codes as well as salient features by twin-projection into a unified model in an adaptive neighborhood-preserving manner. As a result, TP-DPL unifies the salient feature extraction, representation and classification. The twin-incoherence constraint on codes and features can explicitly ensure high intra-class compactness and inter-class separation over them. TP-DPL also integrates the adaptive weighting to preserve the local neighborhood of the coefficients and salient features within each class explicitly. For efficiency, TP-DPL uses Frobenius-norm and abandons the costly $l_0/l_1$-norm for group sparse representation. Another byproduct is that TP-DPL can directly apply the class-specific twin-projective reconstruction residual to compute the label of data. Extensive results on public databases show that TP-DPL can deliver the state-of-the-art performance.

## I. INTRODUCTION

Compact representation learning is an important topic in the communities of data mining and pattern classification for a wide variety of complex data. For the high-dimensional data understanding, lots of compact learning methods have been developed, e.g., subspace learning, matrix factorization and sparse representation. Due to the great success to a variety of real emerging applications, e.g., denoising [14], clustering [8], visual saliency and image classification [1-13][36-40], sparse representation through *Dictionary Learning* (DL) has received much attention in recent years. DL aims to compute the compact representation of data by a linear combination of a few highly correlated atoms in a dictionary [15][41-45]. Thus, the discriminating ability of the dictionary atoms will determine the accuracy of the linear reconstruction over the atoms. It is also worth noticing that the dictionary size, i.e., number of atoms, also has direct effect on the complexity of the compact representation of data. Thus, learning a good dictionary with the strong distinguishing power is crucial for the data representation and classification [1-12][41-45].

One most popular compact DL method is the *K-Singular Value Decomposition* (KSVD) [3]. By the alternates between the sparse coding over dictionary and a process of updating atoms to better fit data, KSVD can represent data effectively, but it does not consider the abilities for discrimination and classification. To use label information for data classification, many discriminative methods were proposed, which can be roughly divided into two kinds, i.e., learning overall shared dictionary of all classes and specific sub-dictionaries over each subject class. The overall shared DL usually applies the discriminative regularization to force the coefficients to be discriminant, among which *Discriminative KSVD* (D-KSVD) [6] and *Label Consistent KSVD* (LC-KSVD) [1] are classical models. D-KSVD adds the classification error into KSVD to enhance classification, while LC-KSVD further adds a label consistency to make the coefficients discriminative.

The category-specific sub-dictionary learning encourages each sub-dictionary to correspond to one single class and the sub-dictionaries of different classes to be as independent as possible. Several representative methods are *Dictionary Learning with Structured Incoherence* (DLSI) [8], *Fisher Discrimination Dictionary Learning* (FDDL) [7], *Projective Dictionary Pair Learning* (DPL) [9], *Structured Analysis Discriminative Dictionary Learning* (ADDL) [11], *Latent Label Consistent Dictionary Learning* (LLC-DL) [10] and *Low-rank Shared Dictionary Learning* (LRSDL) [16]. For representation learning, DPL is clearly different from DLSI, FDDL and LRSDL, since it extends the regular dictionary learning to dictionary pair learning. The analysis dictionary analytically codes data, and the synthesis dictionary is used to reconstruct given data [9]. Based on DPL, ADDL further includes the analysis incoherence promoting function and extends the dictionary pair learning into the joint analysis multi-class classifier training over extracted coefficients. By the analysis DL, DPL and ADDL are efficient for encoding both inside and outside data, but they both cannot encode the salient features of samples jointly to make the reconstruction more accurate for representations. To solve this issue, recent LLC-DL aims at decomposing given data into a latent sparse reconstruction over a structured latent weighted dictionary, a salient feature part and an error part fitting noise. That is, LLC-DL integrates the compact representation and salient feature extraction into a unified model. Although LLC-DL can extract salient features of samples jointly, it suffers from an obvious drawback that it cannot change the dictionary size flexibly. LLC-DL also cannot ensure the high inter-class separation and intra-class compactness of the learnt salient features over different classes, because there is no explicit constraint on the projection for feature extraction. Thus, it would be better to define an explicit incoherence constraint on the sub-projection over each class $l$ so that it computes a salient feature subspace where the training points from class $j\ (j \ne l)$ can be projected to a null space. Another drawback of LLC-DL is that it has to involve an extra time-consuming reconstruction process for each new sample to calculate the coefficient vector, which is in fact also suffered by existing D-KSVD, LC-KSVD, FDDL, DLSI and LRSDL. It is also worth noting that the aforementioned criteria usually cannot preserve the local manifold structures of coding coefficients or (and) salient features, especially in an adaptive manner, which has been proved to be important for improving the descriptive ability in reality. Another potential drawback is that existing methods aim at minimizing the reconstruction error between given data and the product of dictionary and coefficients directly, but such a reconstruction may be too hard and hence causing the overfitting issue in practice.

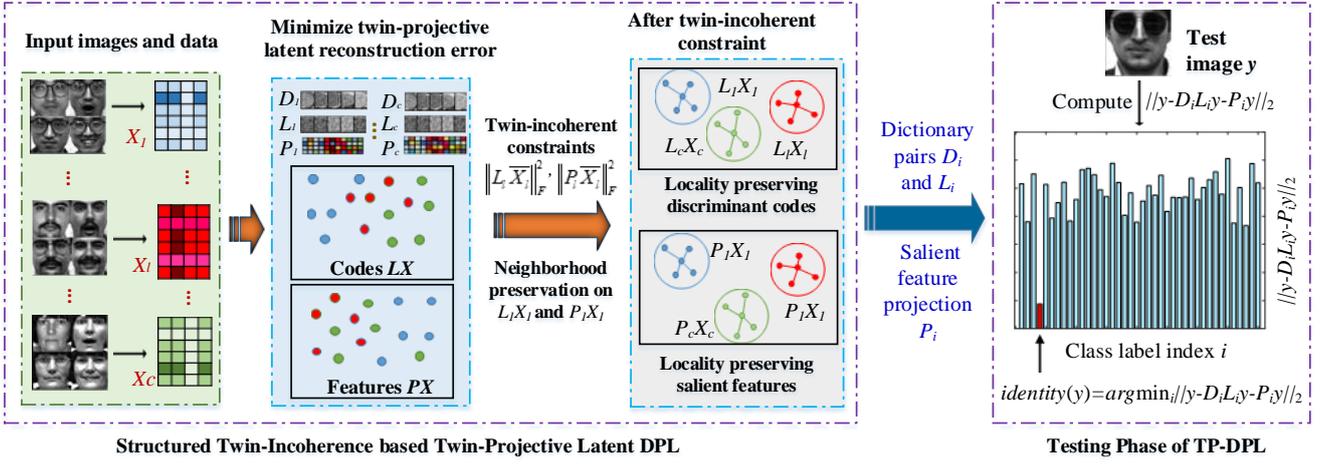

Fig. 1: The schematic flow-diagram of our proposed RA-DPL framework for image representation and recognition.

In this paper, we therefore propose simple and effective strategies to overcome the aforementioned existed problems. The major contributions are described as follows:

(1) A novel structured Twin-Projective Latent Adaptive Flexible DPL (TP-DPL) framework is proposed technically. TP-DPL clearly extends the popular dictionary pair learning into twin-projective latent flexible DPL under the structured twin-incoherence. Specifically, we seamlessly integrate the twin-incoherence based latent adaptive flexible DPL and the joint embedding learning of coefficients as well as salient features by twin-projection into a unified framework. Thus, TP-DPL can involve outside new data efficiency, and will be applicable to the practical applications that need fast online computation. The twin-projective twin-incoherent flexible reconstruction error can avoid the potential over-fitting issue for more accurate reconstruction and representation.

(2) To improve the discriminating abilities of coefficients and salient features by twin-projection, TP-DPL explicitly imposes the twin-incoherence constraint on the coefficients and salient features of samples over each class $l$, which can learn a salient feature subspace where the training samples from class $j\ (j \ne l)$ can be projected to a nearly null space to ensure the intra-class compactness and inter-class separation over them clearly. Moreover, class-specific dictionaries and salient feature are obtained at the same time and the latent sub-dictionaries and salient features over inter-class samples can be independent as much as possible.

(3) To enhance the representation ability by preserving the local manifold structures of both extracted coefficients and salient features during the learning process of structured DL, our TP-DPL clearly integrates the adaptive reconstruction weight learning over learnt coefficients and salient features of each subject class into the twin-projective latent adaptive DPL process. That is, the reconstructive relationship within each class is clearly shared in the sparse representation space and salient feature space, which enables TP-DPL to enhance the representation ability by encoding local neighbourhood.

The paper is outlined as follows. In Section II, we briefly review the related work. Sections III proposes our TP-DPL. Section IV shows the connections with other related models. Section V, we describe the settings and results on the public databases. Finally, the paper is concluded in Section VI.

## II. RELATED WORK

We will briefly introduce the closely related DL approaches.

### A. Review of Overall Dictionary Learning

Let $X = [x_1, \cdots x_i, \cdots x_N] \in \mathbb{R}^{n \times N}$ be a set of training samples from $c$ classes, where $x_i$ is a sample vector, $n$ is the dimension of the original data and $N$ is the number of samples. Then, DL can compute a dictionary $D = [d_1, \cdots d_K] \in \mathbb{R}^{n \times K}$ of $K$ atoms and the coding coefficients matrix $S = [s_1, \cdots s_N] \in \mathbb{R}^{K \times N}$ of $X$ from the following general formulation:

$$\langle D, S \rangle = \arg\min_{D,S} \|X - DS\|_F^2 + \lambda \|S\|_p, \quad (1)$$

where $\|X - DS\|_F^2$ is the reconstruction error over the data $X$, $K$ is the total number of atoms over all classes, and $\lambda > 0$ is a scalar constant. $\|S\|_p$ is a $l_p$-norm regularization, where $p = 0/1$ corresponds to $l_0$-/$l_1$-norm, which is widely-used to ensure the sparse properties of $S$, but such an operation usually incurs a heavy computation burden in reality.

### B. Structured Dictionary Learning by DPL

To avoid the computation burden caused by the costly $l_0$ or $l_1$-norm, DPL was recently derived to calculate a synthesis dictionary $D$ and an analysis dictionary $L$ jointly by solving the following dictionary pair learning problem:

$$\langle L^*, D^* \rangle = \arg\min_{L,D} \sum_{l=1}^{c} \|X_l - D_l L_l X_l\|_F^2 + \lambda \|L_l \overline{X_l}\|_F^2, \ s.t. \|d_i\|_2^2 \le 1, (2)$$

where $\overline{X_l}$ is the complementary data matrix of $X_l \in \mathbb{R}^{n \times N_l}$ in $X$, i.e., excluding $X_l$ itself from $X$. $D_l = [d_1, \cdots d_k] \in \mathbb{R}^{n \times K_l}$ is the synthesis dictionary of class $l$, $L_l \in \mathbb{R}^{K_l \times n}$ is the analysis sub-dictionary of the class $l$, $K_l$ is the number of dictionary atoms according to class $l$, $K$ is the total number of atoms, $L = [L_1; \cdots L_l; \cdots L_c] \in \mathbb{R}^{K \times n}$, $D = [D_1, \cdots D_l, \cdots D_c] \in \mathbb{R}^{n \times K}$ and $N_l$ is the number of samples in class $l$. The constraint $\|d_i\|_2^2 \le 1$ can avoid the trivial solution $L_l = 0$ and make the computation stable. Note that DPL applies Frobenius-norm rather than costly $l_0/l_1$-norm to impose the group sparsity on coefficients matrix $LX$ (i.e., $LX$ is nearly block-diagonal).

In the test phase, if the query sample $y$ is from the class $k$, its projective coding vector by $L_k^*$ will be more likely to be significant, while its projective coding vectors by $L_i^*$, $i \neq k$, tend to be small. Consequently, the reconstruction residual $\|y - D_k^* L_k^* y\|_2^2$ will be much smaller than $\|y - D_i^* L_i^* y\|_2^2$, $i \neq k$. Thus, the class-specific reconstruction residual can be used to identify the label of $y$ by the following criterion:

$$identity(y) = \arg\min_i \|y - D_i L_i y\|_2. \quad (3)$$

That is, the new sample $y$ will be assigned to class $i$ that minimizes the reconstruction error $\|y - D_i L_i y\|_2$.

## III. STRUCTURED TWIN-INCOHERENCE BASED TWIN-PROJECTIVE LATENT ADAPTIVE DPL (TP-DPL)

### A. The Objective Function

To make the representation more accurately and enable the model to extract salient features form given samples jointly, TP-DPL discusses the twin-projective latent flexible DPL problem under the structured twin-incoherence. Specifically, TP-DPL encodes the twin-projective flexibly-relaxed latent reconstruction error $\|X_l + a_l e^T - D_l L_l X_l - P_l X_l\|_F^2$, which avoids the potential over-fitting issue and makes the representation more accurate, where $a_l \in \mathbb{R}^{n \times 1}$ is a bias, $e \in \mathbb{R}^{N_l \times 1}$ is a column vector of all ones, $L_l X_l$ encodes the coefficients of samples in class $l$ and $P_l X_l$ are salient features of $X_l$. That is, our TP-DPL integrates the sparse representation and feature extraction into a unified framework similarly as [10]. The difference between TP-DPL and LLC-DL will be discussed shortly. In addition, a structured twin-incoherence function $r(L_l, P_l)$ on codes $L_l X_l$ and features $P_l X_l$ is used to ensure intra-class compactness and inter-class separation over $L_l X_l$ and $P_l X_l$ jointly. Specifically, if $L_l \in \mathbb{R}^{K_l \times n}$ is the projective analysis latent sub-dictionary and $P_l \in \mathbb{R}^{n \times n}$ is sub-projection corresponding to class $l$, we hope that both $L_l$ and $P_l$ can project the representation and features of training samples of class $j$ ($j \neq l$) to a nearly null space similarly as [9], i.e.,

$$L_l X_j \approx 0, \ P_l X_j \approx 0, \ \forall j \neq l. \quad (4)$$

Thus, the salient features $PX$ will be discriminant in terms of high intra-class compactness and inter-class separation, and the coefficients $LX$ will be nearly block diagonal. Then, we can formulate the following structured twin-incoherence function to improve the discriminating power of TP-DPL:

$$r(L_l, P_l) = \left(\|L_l \overline{X_l}\|_F^2 + \|P_l \overline{X_l}\|_F^2\right), \quad (5)$$

where $\overline{X_l}$ is the complementary data matrix of $X_l$ over $X$. To make the coding efficient, TP-DPL uses the Frobenius-norm and abandon the costly $l_0/l_1$-norm for group sparsity [9-10].

In addition, we also investigate how to preserve the local neighborhood information of both $L_l X_l$ and $P_l X_l$ within each class $l$ clearly in an adaptive manner. Specifically, a structured twin-projection based adaptive reconstruction weighting function $g(L_l, P_l, W_l)$ is involved by computing an adaptive reconstruction weight matrix $W_l \in \mathbb{R}^{N_l \times N_l}$ of each class $l$, where $N_l$ is the number of samples in class $l$. Since we hope that the processes of extracting salient features $P_l X_l$ and learning coefficients $L_l X_l$ can keep neighbourhood information of samples within each class, we minimize the neighborhood reconstruction errors $\|P_l X_l - P_l X_l W_l\|_F^2$ and $\|L_l X_l - L_l X_l W_l\|_F^2$ jointly. Thus, we can define the adaptive reconstruction weighting function as follows:

$$g(L_l, P_l, W_l) = \|P_l X_l - P_l X_l W_l\|_F^2 + \|L_l X_l - L_l X_l W_l\|_F^2 + \|W_l\|_F^2. \quad (6)$$

That is, the weight matrix $W_l$ is shared in the projective feature space spanned by $P_l$ and projective representation space spanned by $L_l$. In addition, the adaptive weighting function $g(L_l, P_l, W_l)$ clearly correlates the coefficients and salient features, i.e., certain important information can be shared in the feature and representation spaces.

By combing Eq.(4), Eq.(5), and Eq.(6), the final objective function of our TP-DPL can be reformulated as

$$\min_{D,L,P,W,a_l} \sum_{l=1}^{c} \Big\{ \|X_l + a_l e^T - D_l L_l X_l - P_l X_l\|_F^2 + \alpha \left(\|L_l \overline{X_l}\|_F^2 + \|P_l \overline{X_l}\|_F^2\right)$$
$$+ \beta \left(\|P_l X_l - P_l X_l W_l\|_F^2 + \|L_l X_l - L_l X_l W_l\|_F^2 + \|W_l\|_F^2\right) \Big\}, \quad (7)$$

where $X_l, D_l, (L_l, P_l)$ and $W_l$ are training data, sub-dictionary, twin-projection pair and adaptive weighting matrix over the class $l$ respectively. In this paper, we omit the constraint $\|d_i\|_2^2 \leq 1$ and the main reason will be described later. $\alpha$ and $\beta$ are two positive weighting factors. Note that the joint minimization of $\alpha r(L_l, P_l) + \beta g(L_l, P_l, W_l)$ can produce the neighborhood preserving discriminative coding coefficients and salient features clearly to enhance the results.

Since the above objective function in Eq. (7) is generally non-convex, we introduce a variable matrix $S$ ($S_l \approx L_l X_l$) to relax the problem as DPL. But note that we apply a flexible term $\|L_l X_l + b_l e^T - S_l\|_F^2$ to avoid the over-fitting and make the representation more accurate, where $b_l \in \mathbb{R}^{K_l \times 1}$ is also a bias, since directly minimizing $\|L_l X_l - S_l\|_F^2$ to approximate $S_l$ using $L_l X_l$ is too hard, which may cause over-fitting. While the term $\|L_l X_l + b_l e^T - S_l\|_F^2$ provides a flexible approximation error clearly. The relaxed optimization problem is

$$\min_{D,L,P,W,S,a_l,b_l} \sum_{l=1}^{c} \Big\{ \|X_l + a_l e^T - D_l S_l - P_l X_l\|_F^2$$
$$+ \alpha \left(\|L_l \overline{X_l}\|_F^2 + \|P_l \overline{X_l}\|_F^2\right) + \gamma \|L_l X_l + b_l e^T - S_l\|_F^2 \quad . \quad (8)$$
$$+ \beta \left(\|P_l X_l - P_l X_l W_l\|_F^2 + \|L_l X_l - L_l X_l W_l\|_F^2 + \|W_l\|_F^2\right) \Big\}$$

Note that the schematic flow-diagram of our framework for image recognition is illustrated in Fig.1, where we show the principles of training and test phases. Next, we detail the optimization procedures of our TP-DPL algorithm.

### B. Optimization

Since the optimization of involved variables depend on each other, the problem cannot be solved directly. Following the common way, we solve the problem by an alternate strategy.

Let $\wp$ be the objective function of our TP-DPL, by taking the derivatives of $\wp$ with respect to $a_l$ and $b_l$, and setting the derivatives to zeros, we can obtain

$$\partial \wp / \partial a_l = a_l e^T e + X_l e - D_l S_l e - P_l X_l e = 0$$
$$\Rightarrow a_l = (D_l S_l e + P_l X_l e - X_l e)/N \quad . \quad (9)$$

$$\frac{\partial \wp}{\partial b_l} = b_l e^T e + L_l X_l e - S_l e = 0 \Rightarrow b_l = (S_l e - L_l X_l e)/N. \quad (10)$$

By the above equations, we can rewrite the flexible errors $\|X_l + a_l e^T - D_l S_l - P_l X_l\|_F^2$ and $\|L_l X_l + b_l e^T - S_l\|_F^2$ as follows:

$$L_l X_l + b_l e^T - S_l = L_l X_l + \left(S_l e e^T - L_l X_l e e^T\right)/N - S_l$$
$$= \left(L_l X_l - L_l X_l e e^T / N\right) - \left(S_l - S_l e e^T / N\right) \quad , \quad (11)$$
$$= L_l X_l H_e - S_l H_e$$

$$\begin{aligned}&X_l + a_l e^T - D_l S_l - P_l X_l \\ &= X_l + \left(D_l S_l ee^T + P_l X_l ee^T - X_l ee^T\right)/N - D_l S_l - P_l X_l \\ &= X_l H_e - D_l S_l H_e - P_l X_l H_e\end{aligned} \quad (12)$$

where $H_e = I - ee^T/N$ is "centering matrix". By substituting the above equations into Eq.(8), we obtain the following equivalent optimization problem for TP-DPL:

$$\min_{D,L,P,W,S} \sum_{l=1}^{c} \left\{ \|X_l H_e - D_l S_l H_e - P_l X_l H_e\|_F^2 \right. \\ + \alpha\left(\|L_l \overline{X_l}\|_F^2 + \|P_l \overline{X_l}\|_F^2\right) + \gamma \|L_l X_l H_e - S_l H_e\|_F^2 \\ \left. + \beta\left(\|P_l X_l - P_l X_l W_l\|_F^2 + \|L_l X_l - L_l X_l W_l\|_F^2 + \|W_l\|_F^2\right)\right\} \quad (13)$$

Then, the above minimization problem can be alternated among the following steps:

*1) Fix P, L, W, update D and S:* By removing the terms irrelevant to $S$ and $D$ from the problem in Eq.(13), we have

$$\langle D,S \rangle = \arg\min_{D,S} \sum_{l=1}^{c} \|X_l H_e - D_l S_l H_e - P_l X_l H_e\|_F^2 \\ + \gamma \|L_l X_l H_e - S_l H_e\|_F^2 \quad (14)$$

Note that the above centered matrix by $H_e$ corresponds to the normalized matrix of original one, which plays a similar role as that of constraint $\|d_i\|_2^2 \leq 1$ in DPL, which can also make the computation of our formulation stable. By taking the derivative w.r.t. $S$ and setting it to zero, we can easily update the coding coefficients $S_l$ of class $l$ as follows:

$$\begin{aligned}S_l &= \left(D_l^T D_l + \gamma I\right)^{-1} A_l \left(H_e H_e^T\right)^{-1} \\ A_l &= D_l^T X_l H_e H_e^T - D_l^T P_l X_l H_e H_e^T + \gamma L_l X_l H_e H_e^T\end{aligned} \quad (15)$$

By taking the derivative w.r.t. $D$ and setting it to zero, we can update the sub-dictionary $D_l$ of class $l$ as

$$D_l = \left(X_l H_e H_e^T S_l^T - P_l X_l H_e H_e^T S_l^T\right)\left(S_l H_e H_e^T S_l^T\right)^{-1} . \quad (16)$$

*2) Fix D, S, W, update P:* By removing terms irrelevant to $P$, we can have the following reduced problem:

$$P = \arg\min_{P} \sum_{l=1}^{c} \|X_l H_e - D_l S_l H_e - P_l X_l H_e\|_F^2 \\ + \alpha \|P_l \overline{X_l}\|_F^2 + \beta \|P_l X_l - P_l X_l W_l\|_F^2 \quad (17)$$

By taking the derivative w.r.t. $P$ and setting it to zero, we can update the projection matrix $P$ as

$$\begin{aligned}P_l &= \left(X_l H_e H_e^T X_l^T - D_l S_l H_e H_e^T X_l^T\right)(B_l + C_l)^{-1} \\ B_l &= X_l H_e H_e^T X_l^T + \beta X_l X_l^T + \alpha \overline{X_l}\,\overline{X_l}^T \\ C_l &= \beta X_l W_l W_l^T X_l^T - \beta X_l W_l X_l^T - \beta X_l W_l^T X_l^T\end{aligned} \quad (18)$$

*3) Fix S and W, update L:* By removing terms irrelevant to $L$ from Eq.(13), we have the following reduced problem:

$$L = \arg\min_{L} \sum_{l=1}^{c} \alpha \|L_l \overline{X_l}\|_F^2 + \gamma \|L_l X_l H_e - S_l H_e\|_F^2 \\ + \beta \|L_l X_l - L_l X_l W_l\|_F^2 \quad (19)$$

By taking the derivative w.r.t. $L$ and setting it to zero, we can update the analysis dictionary $L$ as

$$\begin{aligned}L_l &= \left(\gamma S_l H_e H_e^T X_l^T\right)\left(F_l + \beta X_l W_l W_l^T X_l^T - \beta X_l W_l X_l^T\right)^{-1} \\ F_l &= \beta X_l X_l^T + \gamma X_l H_e H_e^T X_l^T + \alpha \overline{X_l}\,\overline{X_l}^T - \beta X_l W_l^T X_l^T\end{aligned} \quad (20)$$

*4) Fix P and L, optimize W:* By removing terms irrelevant to $W$ from Eq. (13), we have the following reduced problem:

$$W = \arg\min_{W} \sum_{l=1}^{c} \beta\left(\|P_l X_l - P_l X_l W_l\|_F^2 + \|L_l X_l - L_l X_l W_l\|_F^2 + \|W_l\|_F^2\right), \quad (21)$$

By taking the derivative w.r.t. $W$ and setting it to zero, we can update the adaptive weighting matrix $W$ as

$$W_l = \left(X_l^T P_l^T P_l X_l + X_l^T L_l^T L_l X_l + I\right)^{-1}\left(X_l^T P_l^T P_l X_l + X_l^T L_l^T L_l X_l\right). (22)$$

For complete presentation of the method, we summarize the optimization procedures of TP-DPL in Table I, where the iteration will stop when the difference between consecutive objective function values in adjacent iterations is less than 0.0001, and the dictionary $D$ is initialized by applying the identical number of training samples of each class.

| Table I: Optimization procedures of TP-DPL |
| --- |
| **Input:** Training data matrix $X$, class label set $Y$, dictionary size $K$, parameters $\alpha$, $\beta$ and $\gamma$. |
| **Output:** $D$, $L$, $P$, $W$, $S$ |
| **Initialization:** Initialize $P^{(0)}$, $S^{(0)}$ and $L^{(0)}$ to be random matrices with unit F-norm; Initialed $W^{(0)}$ by the cosine similarity between training samples; Initialize $D^{(0)}$ using training samples; $t=0$. |
| **while** *not converge* **do** |
| 1: Update the sparse coefficients $S_l^{(t+1)}$ by Eq.(15); |
| 2: Update the analysis dictionary $D_l^{(t+1)}$ by Eq.(16); |
| 3: Update the salient feature coefficients $P_l^{(t+1)}$ by Eq.(18); |
| 4: Update the synthesis latent dictionary $L_l^{(t+1)}$ by Eq.(20); |
| 5: Update the adaptive weight matrix $W_l^{(t+1)}$ by Eq.(22); |
| 6: $t = t+1$; |
| ***end* while** |

### C. Convergence Analysis

The problem of TP-DPL is solved alternately, so we want to analyze its convergence. Note that TP-DPL is essentially an alternate convex search (ACS) algorithm [20-22], so we can have the following remarks [20-22] to assist the analysis.

**Theorem 1** [22]. If $B \in \mathbb{R}^{n \times m}$, $f : B \to \mathbb{R}$ is bounded and the optimization of variables in each iteration are solvable, the generated sequence $\{f(z_i)\}_{i \in t}$ ($z_i \in B$) by using the ACS algorithm will converge monotonically.

**Theorem 2** [22]. Let $X \subseteq \mathbb{R}^n, Y \subseteq \mathbb{R}^m$ be the closed set and let $f: X \times Y \to \mathbb{R}$ be continuous. Let the optimization of each variable in each iteration be solvable, then we can have: (1) If the sequence $(z_i)_{i \in t}$ by ACS is contained within a compact set, the sequence will contain at least one accumulation point; (2) For each accumulation point $z^*$ of sequence $(z_i)_{i \in t}$: (a) if the optimal solution of one variable with others fixed in each iteration is unique, then all accumulation points will be the local optimal solutions and have the same function value; (b) if the optimal solution of each variable is unique, then we have $\lim_{i \to \infty}\|z_{i+1} - z_i\| = 0$, and the accumulation points can form a compact continuum $C$.

Based on the Theorem 1 and Theorem 2, we can present three remarks on the convergence of our TP-DPL.

**Remark 1.** The generated sequence $\{f(D^i, S^i, P^i, L^i, W^i)\}_{i \in t}$ by our TP-DPL converges monotonically in each iteration.

*Proof.* For the objective function of TP-DPL in Eq.(13), $W$, $P$, $S$, $D$, $L$ are major variables. From the procedures, if $D$, $L$ and $W$ are fixed, the objective function is convex for $S$ and $P$; if $S$ and $P$ are fixed, the function is convex for $D$, $L$ and $W$. In other words, the objective function of TP-DPL is a bi-convex problem for $\{(D,L,W),(S,P)\}$ and the proposed optimization method is actually an alternate convex search

(ACS) algorithm whose convergence analysis has already been intensively studied [20-22]. According to [20-22], the optimal solutions of $(D,L,W)$ and $(S,P)$ can correspond to the iteration steps in ACS, and the problem has a general lower bound 0. As a result, our algorithm is guaranteed to converge to a stationary point in terms of energy.

**Remark 2**. The sequence of $\{D^i,S^i,P^i,L^i,W^i\}_{i\in t}$ generated by our TP-DPL has at least one accumulation point. All the accumulation points are local the optimal solutions of $f$ and moreover have the same function value.

*Proof.* It is easy to check the problem of TP-DPL satisfies $f(D,L,S,P,W)$ for $\|D_l\|_F \to \infty$, $\|S_l\|_F \to \infty$, $\|P_l\|_F \to \infty$ and $\|L_l\|_F \to \infty$. Thus, the generated sequence $\{D^i,S^i,P^i,L^i,W^i\}_{i\in t}$ is bounded in finite dimensional space, and the compact set condition in Theorem 2 (*Condition 1*) is satisfied. Thus, the sequence has at least one accumulation point. By *Theorem 2* (*Condition 2a*), all accumulation points are local optimal and have the same functional value.

**Remark 3**. If $D$, $S$ and $P$ have unique solutions, then the sequence $\{D^i,L^i,S^i,P^i,W^i\}_{i\in t}$ generated by TP-DPL satisfies:

$$\lim_{i\to\infty}\|D^{i+1}-D^i\|+\|L^{i+1}-L^i\|+\|P^{i+1}-P^i\|+\|W^{i+1}-W^i\|=0. \quad (23)$$

*Proof.* Based on *Remark 2*, the *Condition 1* and *2a* in the Theorem 2 are satisfied in TP-DPL, if we have the unique optimal solutions of $L$ and $W$, then we have the conclusion Eq. (23) based on the *Condition 2b* in *Theorem 2* [22]. So, it is easy to check that our TP-DPL is a reasonable approach.

### D. Classification Approach

We discuss the classification approach using TP-DPL. After convergence of TP-DPL, the synthesis dictionary $D_k^*$ and analysis dictionary $L_k^*$ can be obtained to produce small coefficients of samples from classes other than $k$, since they can only generate significant coefficients for the samples of the class $k$. Moreover, the projection $P_k^*$ is also trained to reconstruct the samples of class $k$ to produce the significant salient feature values $P_k^*X_k^*$ and small feature values for the samples of classes other than $k$. As a result, the residual $\|X_k - D_kL_kX_k - P_kX_k\|_2$ will be potentially small over class $k$. On the other hand, since $D_k^*$, $L_k^*$ and $P_k^*$ are not trained to reconstruct $X_i(i\neq k)$, both $D_k^*L_k^*X_i$ and $P_k^*X_i$ are small, i.e., the residual $\|X_i - D_kL_kX_i - P_kX_i\|_2$ will be large.

In the testing phase, if a query sample $y$ is from class $k$, its twin-projective dictionaries $D_k^*$ and $L_k^*$, and salient feature vector by $P_k^*$ will be more likely to be significant, while its latent dictionary vector by $L_i(i\neq k)$ and salient feature coding vector by $P_i(i\neq k)$ tend to be small. As a result, the reconstruction residual $\|y - D_kL_ky - P_ky\|_2$ will be much smaller than residual $\|y - D_iL_iy - P_iy\|_2, i\neq k$. Thus, the class-specific reconstruction residual can be used to identify the label of sample $y$. Thus, similarly as [4][9] we can naturally define the following classifier associated with our TP-DPL:

$$identity(y) = \arg\min_i \|y - D_iL_iy - P_iy\|_2, \quad (24)$$

where $D_l$ is the trained synthesis dictionary of class $l$, $L_l$ is the analysis sub-dictionary, $P_l\in\mathbb{R}^{n\times n}$ is the sub-projection.

## IV. DISCUSSION: RELATIONSHIP ANALYSIS

### A. Connection to the DPL algorithm [9]

The most related model to our TP-DPL is DPL, and we will show that DPL is a special case of our TP-DPL. Recalling the objective function of TP-DPL in Eq.(7), suppose that we constrain $\beta=0$ and $P\to 0$, the problem can be reduced to

$$\min_{D,L,a_l}\sum_{l=1}^c\|X_l + a_le^T - D_lL_lX_l\|_F^2 + \alpha\|L_l\overline{X}_l\|_F^2, \quad s.t. \|d_i\|_2^2\leq 1, \quad (25)$$

which is just the flexibly-relaxed formulation of DPL, since the above problem changes to DPL if $a_l = 0$. But setting $\beta=0$ means that local information of coefficients and salient features cannot be preserved any more and setting $a_l = 0$ to minimize $\|X_l - D_lL_lX_l\|_F^2$ may make the reconstruction from the over-fitting. Thus, our TP-DPL will be superior to DPL.

### B. Connection to the LLC-DL algorithm

We also discuss the connection between our TP-DPL and LLC-DL [10]. To facilitate the comparison, we first present the objective function of LLC-DL as follows:

$$\min_{\substack{D,S,P,\\ \hat{W},A,E}}\sum_{l=1}^c\left\{\|X_l - D\text{diag}(V_l)S_l - PX_l\|_F^2 + \alpha\left[\|S_l\|_1 + \|\hat{Q}_l - A_lS_l\|_F^2\right]\right\}$$
$$+\lambda\sum_{l=1}^c\sum_{q\neq l}\sum_{p=1}^K\sum_{r\neq p}v_{l,r}(d_r^Td_p)^2 v_{q,p} + \beta\left[\|H - \hat{W}PX\|_2^2 + \|\hat{W}^T\|_{2,1}\right], \quad (26)$$

where $H$ is class label set and $\|\hat{W}^T\|_{2,1}$ is the $l_{2,1}$-norm based robust classifier. By comparing the above problem with our TP-DPL, we can find that: (1) LLC-DL discusses latent DL under a structured discriminative sparse code error, while we discuss the twin-projective latent flexible adaptive learning of dictionary pairs (a synthesis dictionary $D$ and an analysis dictionary $L$) under a structured twin-incoherence. Moreover, the function of the adaptive weighting function $g(L_l,P_l,W_l)$ in TP-DPL correlates the coefficients and salient features, while LLC-DL cannot mine shared vital information in the feature and representation spaces; (2) For the salient feature extraction, LLC-DL learns a shared projection $P$ for all the classes, while TP-DPL learns sub-projections $P_l$ for various classes. More importantly, TP-DPL imposes an incoherence constraint on salient features over different classes to ensure high inter-class separation by encouraging each $L_l$ to project the training samples of class $j$ ( $j \neq l$ ) to a nearly null space. But LLC-DL cannot ensure this issue clearly; (3) LLC-DL computes the representation matrix $S_l$ of each class directly, so it needs the extra time-consuming sparse reconstruction process with well-trained dictionary $D$ to obtain the codes of each new data. In contrast, our TP-DPL learns an analysis dictionary $L$ jointly, which can be used to extract the codes from inside and outside data efficiently. In addition, TP-DPL applies the Frobenius-norm for preserving the group sparse properties similarly as DPL [9], while LLC-DL applies the $l_1$-norm on the coefficients, but the optimization of $l_1$-norm is usually time-consuming; (4) LLC-DL incorporates the classification error over the extracted features for the joint optimization by involving an extra tuning parameter, but the optimal parameter selection is usually difficult in reality. While our TP-DPL will not suffer from this issue, since our method minimizes the reconstruction residual to determine the label of each outside new data directly.

## V. EXPERIMENTAL RESULTS AND ANALYSIS

We mainly evaluate our TP-DPL for data representation and classification, and the performance of TP-DPL is compared with those of closely related SRC [4], DLSI [8], KSVD [3], D-KSVD [6], LC-KSVD [1], COPAR [32], FDDL [7], DPL [9], ADDL [11], LLC-DL[10] and LRSDL [16]. Since DLSI and KSVD did not define an explicit classification method, we apply the same classification approach of SRC for them.

Five face databases (i.e., ORL [25], YaleB [26], UMIST

[27], AR [28] and CMU PIE [29]), an object database (i.e., ETH80 [30]) and a scene image database (i.e., the fifteen scene categories database [31]) and are used for evaluations. Note that these datasets are widely-used to test the results of DL [1-11]. The details of the datasets are described in Table II, where we report the numbers of samples, dimensions and subjects. Note that we follow the common procedures and resize all images of ORL, AR, YaleB, CMU PIE, UMIST and ETH80 into 32×32 pixels, so each image corresponds to a point in a 1024-D space. For classification, we randomly split each database into a training set and a test set. For fair comparison, the accuracy is averaged over 15 random splits of training/test samples to avoid the bias by the randomness. We perform all the simulations on a PC with Intel (R) Core (TM) i7-7700 CPU @ 3.6 GHz 8G.

TABLE II.
DETAILED DESCRIPTIONS OF USED REAL DATABASES.

| Dataset Name | # Samples | # Dim | # Classes |
|---|---|---|---|
| ORL face | 400 | 1024 | 40 |
| YaleB face | 2414 | 504 | 38 |
| AR face | 2600 | 540 | 100 |
| CMU PIE face | 11554 | 1024 | 68 |
| UMIST face | 1012 | 1024 | 20 |
| ETH80 object | 3280 | 1024 | 80 |
| Fifteen scene categories | 4485 | 3000 | 15 |

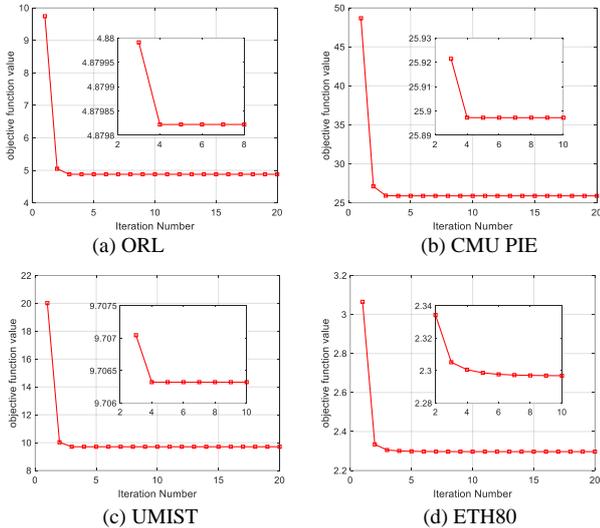

**Fig. 2:** Convergence behavior of TP-DPL, where the *x*-axis is the number of iterations and the *y*-axis is the objective function value.

### A. Convergence Analysis

We mainly analyze the convergence behavior by describing the objective function value in this experiment. ORL, CMU PIE, UMIST and ETH80 are used [11]. For ORL, CMU PIE, UMIST, and ETH80 object database, we randomly select 5, 30, 10, 6 images from each subject for training respectively, and set the dictionary size as the number of training samples. The averaged results over 20 iterations are shown in Fig.2. We find that the objective function value of our TP-DPL is non-increasing in the iterations and the number of iterations is usually less than 10 in most cases.

### B. Parameter Selection Analysis

The parameter selection issue still remains an open problem, thus we apply a heuristic way to select the most important parameters. Note that our TP-DPL has three parameters (i.e., $\alpha, \beta$ and $\gamma$), so we fix one of the parameters and explore the effects of other two on the performance by the grid search strategy. AR face and ETH80 object databases are used as examples. For AR, we use the convolutional features and 20 images in each individual are randomly chosen for training and the number of atoms is set to 100, corresponding to an average of 5 items per person. For ETH80, we follow [11] to use discriminant features [33], select 6 images from each class for training, test on the rest and select the number of atoms corresponding to an average of 6 items per class. For each pair of parameters, we average the results over varied parameters from $\{5\times10^{-5}, 5\times10^{-4}, 5\times10^{-3}, 5\times10^{-2}, 5\times10^{-1}, 5, 5\times10^{1}, 5\times10^{2}, 5\times10^{3}, 5\times10^{4}, 5\times10^{5}\}$. The parameter selection results are shown in Fig.3. As can be seen, our TP-DPL performs well in a wide range of parameters in each group, i.e., it is insensitive to the model parameters.

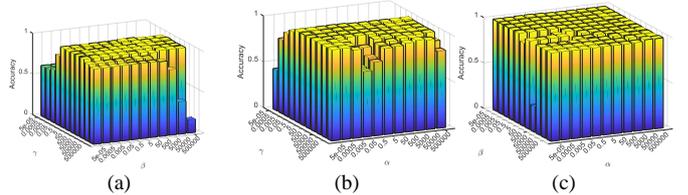

**Fig. 3-1:** Parameter sensitivity analysis of TP-DPL on AR, where (a) tune $\gamma$ and $\beta$ by fixing $\alpha$=50 ; (b) tune $\gamma$ and $\alpha$ by fixing $\beta$=0.5 ; (c) tune $\alpha$ and $\beta$ on the performance by fixing $\gamma$=0.5 .

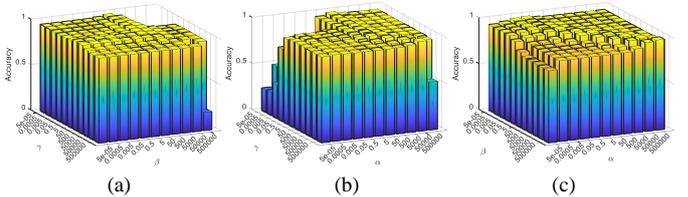

**Fig. 3-2:** Parameter sensitivity analysis of TP-DPL on ETH80, where (a) tune $\gamma$ and $\beta$ by fixing $\alpha$=50 ; (b) tune $\gamma$ and $\alpha$ by fixing $\beta$=50 ; (c) tune $\alpha$ and $\beta$ on the performance by fixing $\gamma$=0.05 .

### C. Application to Image Recognition

We evaluate each method for representing and recognizing three kinds of image databases, i.e., face images (i.e., YaleB, AR, CMU PIE, and UMIST), ETH80 object database, and the fifteen nature scene categories database. Some image examples of these databases are shown in Fig.4. For each method, we choose the model parameters carefully. Since KSVD, D-KSVD and LC-KSVD applies the $l_0$-norm based sparsity constraint for DL, we still use the $l_0$-norm for them for fair comparison. The averaged recognition results are reported as the evaluation metric of each algorithm.

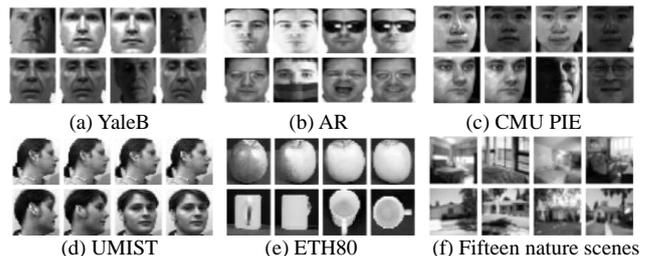

(a) YaleB (b) AR (c) CMU PIE
(d) UMIST (e) ETH80 (f) Fifteen nature scenes
**Fig. 4:** Sample images of the evaluated real-world databases.

**Face Recognition on YaleB.** we use random face features [1-3][13][18-19] in this study, i.e., each image is projected onto a 504-dimensional vector by a generated matrix from a zero-mean normal distribution, and each row of matrix is $l_2$ normalized. We clearly follow the setting in [9] for the fair comparison, i.e., half of the images per class are randomly selected for training and the rests are used for testing. The dictionary contains 570 items, corresponding to an average of 15 items of each class. The averaged recognition rates are reported in Table III, where $\alpha$=0.0005, $\beta$ = 500 and $\lambda$ = 0.5 are set in TP-DPL. The results of other compared methods are adopted directly from [9]. We find that TP-DPL delivers higher accuracy than its competitors under the same setting.

**Face Recognition on AR.** In this study, by following the common procedures [1-3][12-14], the face set that contains 2600 images of 50 males and 50 females is evaluated. We clearly follow [1-3][13][19] to use 540-dimensional random face features. We also randomly choose 20 images from each person for training and test on the rest. The dictionary contains 500 items, corresponding to an average of 5 items per category. $\alpha=0.0005$, $\beta=50000$ and $\gamma=0.5$ are used for our TP-DPL. The results are described in Table IV, where the results of compared methods are adopted from [1][11]. We find from the results that our TP-DPL can deliver enhanced results than its competitors under the same setting.

TABLE III.
RECOGNITION RESULTS USING RANDOM FACE FEATURES ON YALEB.

| Evaluated Methods | Accuracy |
|---|---|
| SRC(all train. sample) | 96.5% |
| K-SVD(15 items) | 93.1% |
| D-KSVD(15 items) | 94.1% |
| LC-KSVD1(15 items) | 94.5% |
| LC-KSVD2(15 items) | 95.0% |
| DLSI (15 items) | 97.0% |
| COPAR (15 items) | 96.9% |
| FDDL(15 items) | 96.7% |
| DPL(15 items) | 97.5% |
| LRSDL(15 items) | 97.3% |
| ADDL(15 items) | 97.6% |
| **Our TP-DPL(15 items)** | **98.2%** |

TABLE IV.
RECOGNITION RESULTS USING RANDOM FACE FEATURES ON AR.

| Evaluated Methods | Accuracy |
|---|---|
| SRC (5 items, 20 labels) | 66.5% |
| KSVD(5 items, 20 labels) | 86.5% |
| D-KSVD(5 items, 20 labels) | 88.8% |
| LC-KSVD1(5 items, 20 labels) | 92.5% |
| LC-KSVD2(5 items, 20 labels) | 93.7% |
| DLSI(5 items, 20 labels) | 93.1% |
| FDDL(5 items, 20 labels) | 95.6% |
| DPL(5 items, 20 labels) | 95.8% |
| LRSDL(5 items, 20 labels) | 96.8% |
| ADDL(5 items, 20 labels) | 97.0% |
| **Our TP-DPL(5 items, 20 labels)** | **98.6%** |

**Face Recognition on CMU PIE.** This database contains 68 persons with 41368 face images as a whole. Follow the common procedures in [2][29], 170 near frontal images per person are employed for the evaluations. This face subset consists of five near frontal pose (C05, C07, C09, and C29) and all the images have different illuminations, lighting and expression. We also adopt random face features [4][17] and set the dimension to 256. We train on 20, 30, and 40 images per person and test on the rest, and set the dictionary size to the number of training images in each study. The averaged results are described in Table V, where $\alpha=50$, $\beta=500$ and $\gamma=0.005$ are used in TP-DPL. We find that: (1) the accuracy increases as the training number increases; (2) our TP-DPL is superior to its competitors in investigated cases.

**Face Recognition on UMIST.** In this study, we randomly select 5 images per class for training and use other images for testing. The number of atoms is set to be the number of training images and we normalize each sample to be the unit $l_2$-norm. $\alpha=0.0005$, $\beta=50$ and $\gamma=0.0005$ are set in TP-DPL. The averaged rates are shown in Table VI. Our TP-DPL can still deliver higher accuracy than other competitors.

TABLE V.
RECOGNITION RESULTS USING RANDOM FACE FEATURES ON CMU PIE.

| Evaluated Methods | 20 train | 30 train | 40 train |
|---|---|---|---|
| SRC | 77.4% | 82.6% | 83.5% |
| KSVD | 78.9% | 83.0% | 84.3% |
| D-KSVD | 80.2% | 83.5% | 85.9% |
| LC-KSVD1 | 81.3% | 85.0% | 87.1% |
| LC-KSVD2 | 81.5% | 85.9% | 87.2% |
| DLSI | 78.3% | 84.5% | 89.1% |
| COPAR | 86.1% | 89.1% | 90.9% |
| FDDL | 84.7% | 89.5% | 91.2% |
| DPL | 86.5% | 89.4% | 90.3% |
| LLC-DL | 87.0% | 89.5% | 90.4% |
| LRSDL | 87.1% | 89.5% | 91.2% |
| ADDL | 87.0% | 89.6% | 91.5% |
| **Our TP-DPL** | **92.3%** | **94.7%** | **95.3%** |

TABLE VI.
RECOGNITION RESULTS ON UMIST FACE DATABASE.

| Evaluated Methods | Mean ± Std(%) |
|---|---|
| SRC (5 items, 5 labels) | 87.4 ± 2.4 |
| KSVD(5 items, 5 labels) | 87.7 ± 2.5 |
| D-KSVD(5 items, 5 labels) | 87.2 ± 2.1 |
| LC-KSVD1(5 items, 5 labels) | 87.8 ± 2.7 |
| LC-KSVD2(5 items, 5 labels) | 88.6 ± 2.0 |
| DLSI(5 items, 5 labels) | 87.1 ± 2.1 |
| COPAR(5 items, 5 labels) | 87.3 ± 2.0 |
| FDDL(5 items, 5 labels) | 87.5 ± 1.6 |
| DPL(5 items, 5 labels) | 88.9 ± 1.6 |
| LLC-DL(5 items, 5 labels) | 89.2 ± 2.0 |
| LRSDL(5 items, 5 labels) | 90.4 ± 2.3 |
| ADDL(5 items, 5 labels) | 90.9 ± 1.7 |
| **Our TP-DPL(5 items, 5 labels)** | **95.0 ± 0.9** |

TABLE VII.
RECOGNITION RESULTS USING SPATIAL FEATURES ON THE FIFTEEN SCENE CATEGORY DATABASE.

| Evaluated Methods | Accuracy |
|---|---|
| SRC (30 items, 100 labels) | 91.8% |
| KSVD(30 items, 100 labels) | 86.7% |
| DKSVD(30 items, 100 labels) | 89.1% |
| LC-KSVD1(30 items, 100 labels) | 90.4% |
| LC-KSVD2(30 items, 100 labels) | 92.9% |
| DLSI(30 items, 100 labels) | 92.5% |
| COPAR(30 items, 100 labels) | 92.9% |
| FDDL(30 items, 100 labels) | 93.1% |
| DPL(30 items, 100 labels) | 96.9% |
| LRSDL(30 items, 100 labels) | 97.1% |
| ADDL(30 items, 100 labels) | 98.1% |
| **Our TP-DPL(30 items, 100 labels)** | **98.8%** |

**Scene Recognition on fifteen categories database.** The nature scene categories database includes fifteen scenes, i.e., suburb, open country, mountain, coast, forest, store, kitchen, office, industrial, living room, tall building, bedroom, street, highway and inside city. Each category contains 200 to 400 images, and each scene image has about $250 \times 300$ pixels. By following [1][9][11], the spatial pyramid features by using a four-level spatial pyramid and a SIFT-descriptor codebook with size 200 are computed for the simulations. The final spatial pyramid features are reduced to 3000 by using PCA [34]. Following the settings in [1][11], we select 100 images per category for training and test on the rest. The dictionary size is set to 450 items, corresponding to an average of 30 items for each category. $\alpha=5\times10^{-5}$, $\beta=5000$ and $\gamma=0.5$ are used in TP-DPL. We describe the averaged results in Table VII, where directly adopt the results of compared methods from [1][11]. We can find that TP-DPL obtains better results than other models under the same experimental setting.

**Object Recognition on ETH80.** ETH80 object database has 3280 images of 80 subcategories from 8 big categories

[30]. Each big category contains 10 subcategories, each of which has 41 images. We follow [11] to use the discriminant features [33]. We similarly select 6 images from each class for training and test on the rest. $\alpha=50$, $\beta=50$ and $\gamma=0.05$ are used in TP-DPL. We show the averaged results in Table VIII. We find that TP-DPL achieves the enhanced result than other evaluated models. DPL, LLC-DL, LRSDL and ADDL also deliver promising results that are highly comparative with TP-DPL. In addition, we also evaluate the recognition rates for individual classes and show some image examples in the eight classes having high accuracy rate in Fig.5.

TABLE VIII.
RECOGNITION RESULTS ON THE ETH80 OBJECT DATABASE

| Evaluated Methods | Mean ± Std (%) |
|---|---|
| SRC (6 items, 6 labels) | 89.6 ± 0.8 |
| KSVD(6 items, 6 labels) | 91.2 ± 0.8 |
| D-KSVD(6 items, 6 labels) | 91.2 ± 0.4 |
| LC-KSVD1(6 items, 6 labels) | 90.7 ± 0.7 |
| LC-KSVD2(6 items, 6 labels) | 91.5 ± 0.8 |
| DLSI(6 items, 6 labels) | 92.7 ± 0.9 |
| COPAR(6 items, 6 labels) | 93.1 ± 0.7 |
| FDDL(6 items, 6 labels) | 93.2 ± 0.3 |
| DPL(6 items, 6 labels) | 97.7 ± 0.2 |
| LLC-DL(6 items, 6 labels) | 97.6 ± 0.2 |
| ADDL(6 items, 6 labels) | 97.9 ± 0.2 |
| LRSDL(6 items, 6 labels) | 97.7 ± 0.2 |
| **Our TP-DPL(6 items, 6 labels)** | **98.3 ± 0.2** |

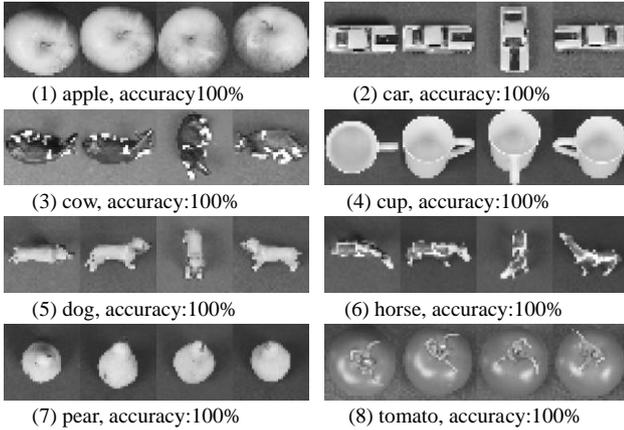

(1) apple, accuracy100%   (2) car, accuracy:100%
(3) cow, accuracy:100%   (4) cup, accuracy:100%
(5) dog, accuracy:100%   (6) horse, accuracy:100%
(7) pear, accuracy:100%  (8) tomato, accuracy:100%

**Fig. 5:** Image examples from the classes with high accuracy rate from the ETH80 object database.

### D. Image Clustering using Convolutional Features

We evaluate each algorithm for clustering and visualization based on the convolutional features of LeNet-5 [35][38]. It is worth noting that our TP-DPL performs twin-projective flexible dictionary pair learning through minimizing a latent reconstruction error among the original samples in *X*, sparse reconstruction *DLX* and the salient features *PX*. Thus, we would like to investigate how the reconstruction *DLX* and salient features *PX* affect the data representation results by clustering analysis and visualization. Two face databases, i.e., AR and CMU PIE, are employed. For the quantitative clustering evaluations, we use the clustering Accuracy (AC) [23-24]. The values of AC are distributed from 0 to 1. The higher the value is, the better the clustering result is. In this study, we divide the convolutional features of each database randomly into a training feature set and a test feature set *Y*. The training set is used for DL to obtain the twin-projective dictionaries *D* and *L* that can then be used to obtain the reconstruction *DLY* and salient features *PY* of the feature set *Y*. We aim at comparing the results by performing k-means clustering with the squared Euclidean distance on *DLY*, *PY* and *DLY +PY* respectively. The number k of clusters on the test set is equal to the number of data categories. For each setting, we average the numerical clustering results over 10 random initializations for the k-means clustering algorithm.

**Results on AR.** In this study, we randomly choose 5, 10, 15 and 20 images from each person for training and test on the rest. The number of dictionary atoms is set to be the size of the training set for our TP-DPL, which corresponds to an average of 5, 10, 15 and 20 items from each category. $\alpha=50$, $\beta=50$ and $\gamma=0.5$ are used in our TP-DPL. We report the mean clustering AC of different runs in Table IX. We find that the clustering result on *DLY+PY* is obviously superior to those on *Y*, *DLY* and *PY*, i.e., *Y*, *DLY* or *PY* fails to capture all important features. In other words, *DLY+PY* can describe given data better. As a result, existing models that directly apply the reconstruction *DS*, coefficients or salient features for data classification or clustering may produce inaccurate results. On the contrary, our TP-DPL minimizes a structured twin-incoherence based twin-projective flexible latent reconstruction error for representation learning and data classification, which is potentially more reasonable.

TABLE IX
CLUSTERING AC USING CONVOLUTIONAL FEATURES ON AR.

| Atom number | Y | PY | DLY | DLY+PY |
|---|---|---|---|---|
| 5(per class) | 71.55% | 51.77% | 51.17% | **86.63%** |
| 10(per class) | 70.11% | 36.68% | 35.74% | **87.51%** |
| 15(per class) | 65.65% | 30.08% | 30.41% | **88.44%** |
| 20(per class) | 60.47% | 32.52% | 31.43% | **89.22%** |

TABLE X
CLUSTERING AC USING CONVOLUTIONAL FEATURES ON CMU PIE.

| Train number | Y | PY | DLY | DLY+PY |
|---|---|---|---|---|
| 10(per class) | 81.66% | 36.08% | 35.96% | **84.47%** |
| 15(per class) | 81.62% | 31.12% | 31.03% | **85.53%** |
| 20(per class) | 80.85% | 31.32% | 30.87% | **86.42%** |
| 25(per class) | 81.03% | 29.59% | 28.89% | **86.60%** |
| 30(per class) | 80.86% | 25.52% | 24.67% | **87.74%** |
| 35(per class) | 81.45% | 23.05% | 23.96% | **88.28%** |
| 40(per class) | 81.87% | 23.05% | 22.35% | **88.78%** |

**Results on CMU PIE face database.** For this database, we randomly choose 10, 15, 20, 25, 30, 35 and 40 images per person for training and test on the rest. The number of dictionary atoms is set to be the size of training set for our TP-DPL. $\alpha=50$, $\beta=50$ and $\gamma=0.005$ are set in TP-DPL. The averaged clustering results are shown in Table X. We can find that the clustering AC on *DLY+PY* can still deliver the enhanced performance over *DLY* or *PY* based on different numbers of training samples, which can once again prove that *DLY+PY* is able to describe the original data better than *DLY* or *PY*. It can also be found that the increasing number of training samples can improve the result on *DLY+PY*.

**Visualization of features on CMU PIE.** We also aim at visualizing the feature set *Y* and *DLY+PY* for observation. Based on the convolutional features of database, we choose 40 images per person for training and use the rest to form *Y*. We simply set the number of atoms to the number of training samples. To better compare the original feature set *Y* and our *DLY+PY* features, the top-10 class are selected from the test set for this visualization task based on the clustering results. Note that we aim to visualize the first 9-dimensions of the features *Y* and *DLY+PY* in Fig.6, from which we observe that the features *DLY+PY* by our TP-DPL are clearly better than convolutional features *Y* by obtaining high intra-class compactness and high inter-class separation, which is a very good message for the feature representation, data clustering and classification in reality.

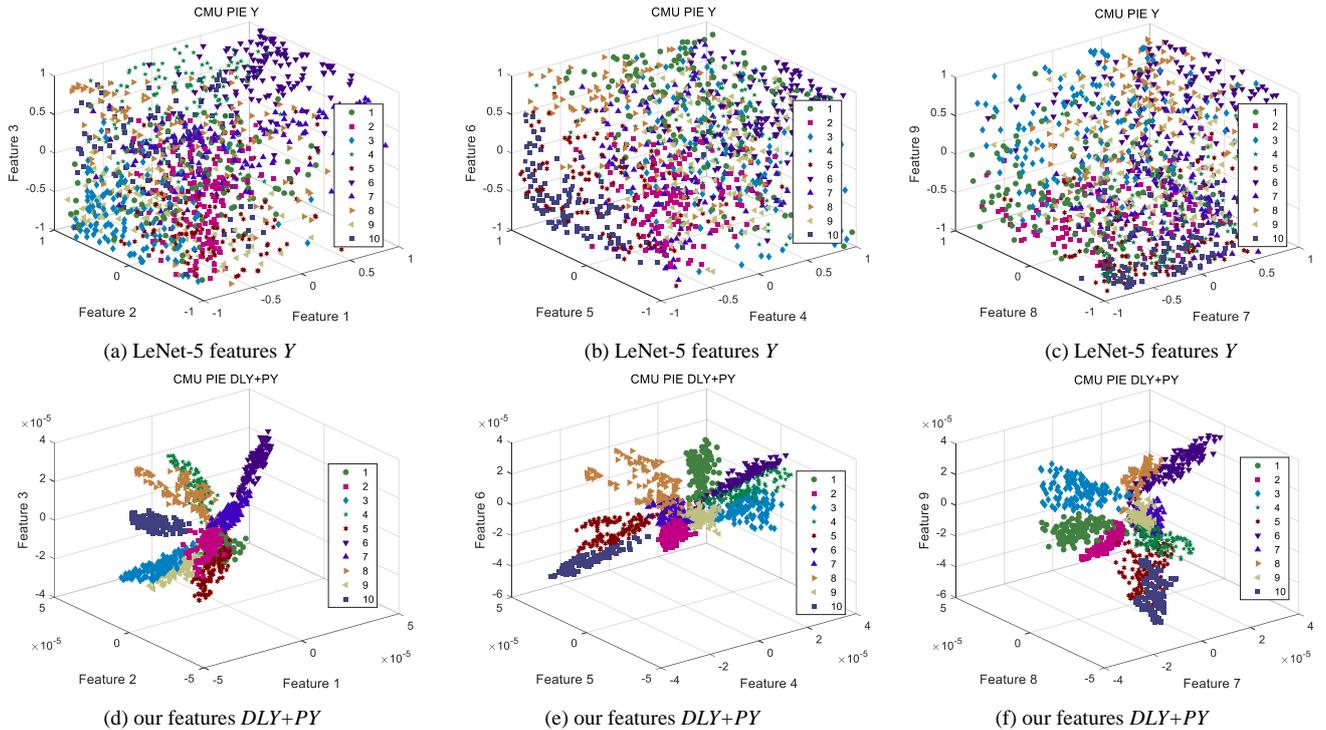

**Fig. 6:** Visualizations of the convolutional features *Y* and features *DLY+PY* by our TP-DPL on CMU PIE, where: (a) 1, 2 and 3-dimensional features of *Y* feature; (b) 4, 5 and 6-dimensional features of *Y*; (c) 7, 8 and 9-dimensional features of *Y*; (d) 1, 2 and 3-dimensional features of *DLY+PY*, (e) 4, 5 and 6-dimensional features of *DLY+PY*; (f) 7, 8 and 9-dimensional features of *DLY+PY*.

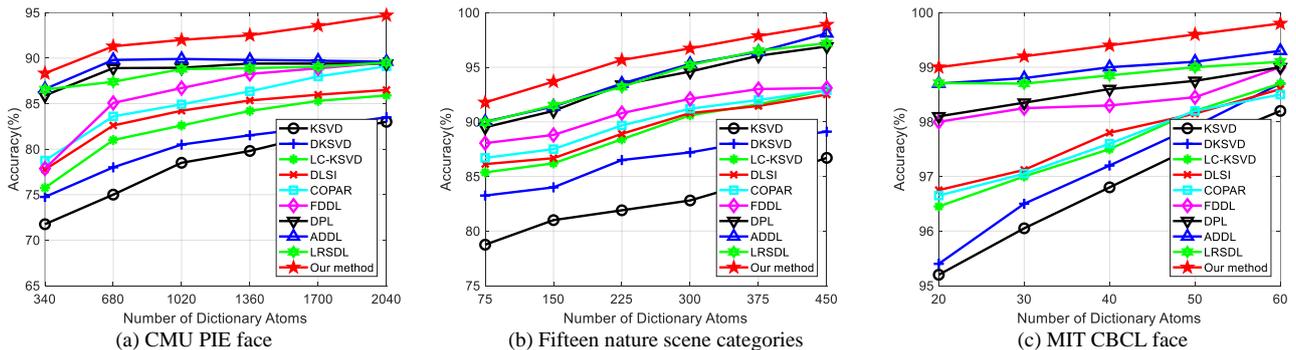

**Fig. 7:** Quantitative recognition evaluation result of each algorithm vs. varying dictionary sizes on three real image databases.

*E. Quantitative Evaluation of Dictionaries*

We mainly evaluate the performance of learnt dictionary *D* of each method. We show the quantitative evaluation results of recognition against varying dictionary sizes. Three image databases, i.e., MIT CBCL, CMU PIE and fifteen nature scene categories databases are evaluated. For CMU PIE, we still use random face features of dimension 256, choose 30 samples per class for training and evaluate each method with varying sizes *K* of dictionary, i.e., *K*=340, 680, 1020, 1360, 1700 and 2040 in Fig. 7a. For fifteen scene categories, we also follow [1][5] to choose 100 samples per class for training and evaluate each method with varying dictionary sizes *K*, i.e., *K*=75, 150, 225, 300, 375 and 450 in Fig.7b. For MIT CBCL, we choose 6 samples per class as training set, test on the rest, evaluate each model with varying sizes *K* of the dictionary, i.e., *K*=20, 30, 40, 50 and 60, and the averaged recognition results are illustrated in Fig.7c. We can find that: (1) the recognition accuracy of each method can be increased when the number of atoms increases; (2) TP-DPL obtains better results than its competitors across all dictionary sizes. DPL, ADDL and LRSDL can also perform well by delivering higher accuracies than other remaining methods. KSVD is the worst method in most cases.

## VI. CONCLUDING REMARKS

We proposed a structured twin-incoherence constrained twin-projective latent adaptive DPL model for classification. TP-DPL unifies the salient feature extraction, representation and classification in an adaptive locality-preserving manner via minimizing a twin-incoherent flexible reconstruction error. The twin-incoherence constraints on coefficients and salient features can produce discriminative coefficients and features with high inter-class separation and high intra-class compactness, which is benefit for enhancing classification.

We evaluated the effectiveness of our method on several public databases for data classification and clustering. The obtained results demonstrated the superior performance of our model. The clustering and visualization of features also verified the effectiveness of the twin-incoherence based twin-projective latent reconstruction to deliver the salient features jointly. In future, we will investigate how to extend our model to the semi-supervised scenario [12] to handle the cases that the number of labeled samples is limited to enhance the latent dictionary pair learning. Extending our method to the other application areas, such as the content or text based image retrieval, is also interesting.

ACKNOWLEDGMENT

This work is partially supported by the National Natural Science Foundation of China (61672365, 61732008, 61725203, 61622305, 61871444, 61572339), High-Level Talent of the "Six Talent Peak" Project of Jiangsu Province of China (XYDXX-055), and the Fundamental Research Funds for the Central Universities of China (JZ2019HGPA-0102). Dr. Zhao Zhang is the corresponding author.